\pdfoutput=1
\documentclass[runningheads]{llncs}
\usepackage{graphicx}
\usepackage{amsmath,amssymb} 
\usepackage{color}
\usepackage[width=122mm,left=12mm,paperwidth=146mm,height=193mm,top=12mm,paperheight=217mm]{geometry}
\usepackage{hyperref}
\hypersetup{
    colorlinks=true,
    linkcolor=blue,
    filecolor=magenta,      
    urlcolor=cyan,
}
\begin{document}
\pagestyle{headings}
\mainmatter
\def\ECCV18SubNumber{2072}  

\title{Aligning Across Large Gaps in Time} 

\titlerunning{Aligning Across Large Gaps in Time}

\authorrunning{Hunter Goforth}

\author{Hunter Goforth, Simon Lucey}
\institute{The Robotics Institute, Carnegie Mellon University}

\newcommand{\tb}[1]{\textbf{#1}}

\maketitle

\begin{abstract}
We present a method of temporally-invariant image registration for outdoor scenes, with invariance across time of day, across seasonal variations, and across decade-long periods, for low- and high-texture scenes. Our method can be useful for applications in remote sensing, GPS-denied UAV localization, 3D reconstruction, and many others. Our method leverages a recently proposed approach to image registration, where fully-convolutional neural networks are used to create feature maps which can be registered using the Inverse-Composition Lucas-Kanade algorithm (ICLK). We show that invariance that is learned from satellite imagery can be transferable to time-lapse data captured by webcams mounted on buildings near ground-level.

\end{abstract}

\section{Introduction}

Image alignment is one of the most fundamental tasks in computer vision, with applications such as object tracking~\cite{YWu}, video stabilization~\cite{SBattiato}, and visual odometry~\cite{JEngel}. In many applications the images to be aligned were taken several milliseconds apart, as in the case of visual odometry or object tracking. In other applications, the images were taken several hours or days apart. This can be the case in robotic localization, where a robot compares its surroundings to a recently created map of the environment~\cite{AMurillo}. Sometimes, applications require alignment across temporal differences on the scale of months, years or decades. This is the case in the remote sensing community, where algorithms have been developed to automatically align satellite images taken at different times of the year, or across years~\cite{YBentoutou,KYang}. This is also the case for 3D reconstruction algorithms which utilize large amounts of imagery taken across months or years~\cite{SAgarwal}.

Alignment algorithms must overcome changes in brightness, illumination, exposure, and geometric warping. In the case of images of outdoor environments, algorithms must also combat with occlusions, seasonal changes, time of day, and the addition or removal of objects in parts of the image like buildings and cars. All of these issues are exacerbated with larger temporal differences between the capturing of the images.

\begin{figure}
\centering
\includegraphics[height=4.5cm]{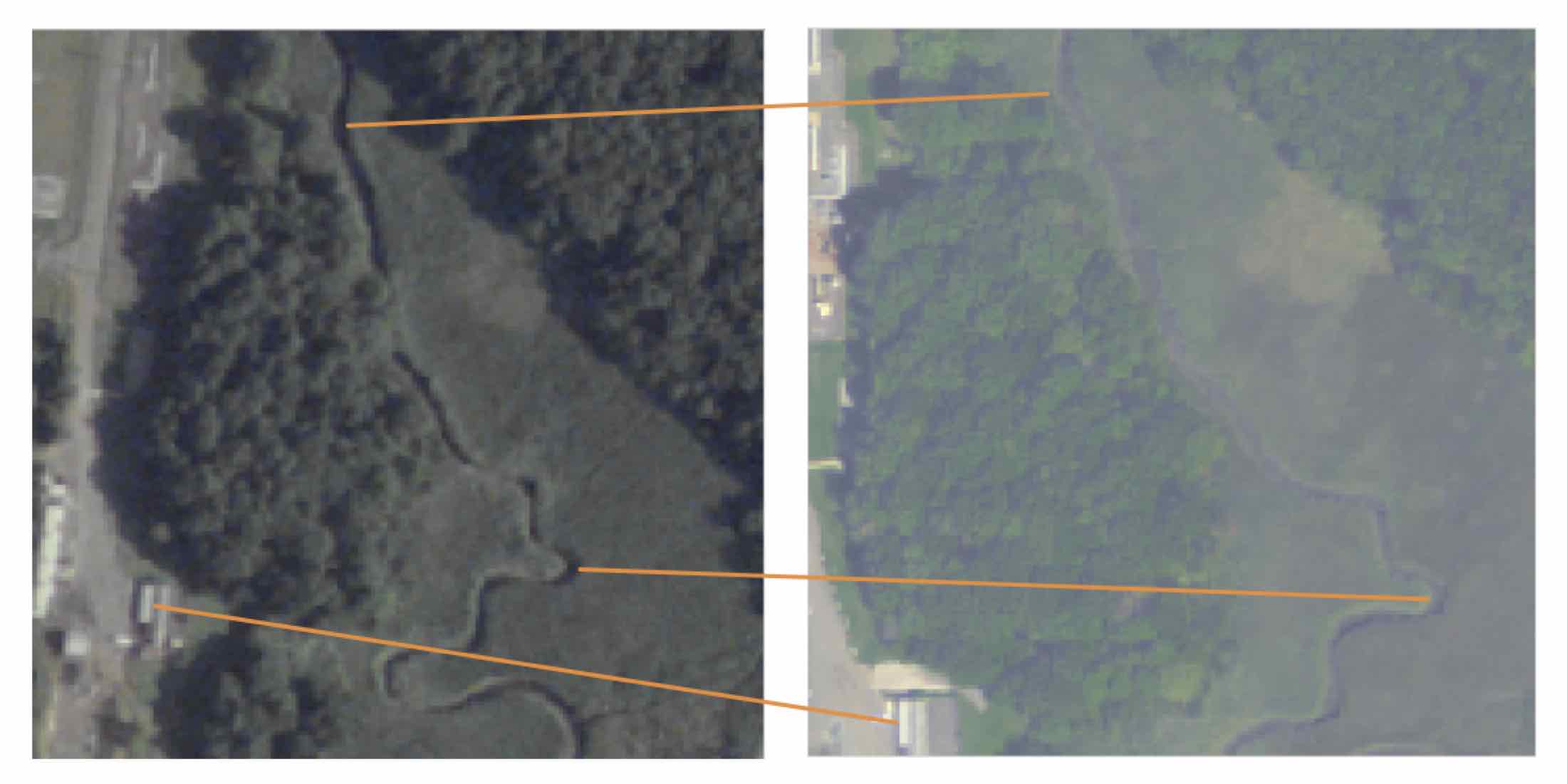}
\caption{An example of images of outdoor environments, taken three years apart and during different seasons. Aligning these images is difficult due to lack of texture, photometric differences, and stylistic differences in the scene like the color of the grass. Overcoming this challenge would be valuable in applications such as GPS-denied UAV localization, where the UAV could register imagery from a downward-facing camera to a pre-captured satellite image. Our paper presents a method capable of performing this alignment.}
\label{example}
\end{figure}

These issues often do not pose a problem for humans when aligning two images. Take, for instance, the two images shown in Figure \ref{example}. One image is taken in summer, and the other in winter. Manually aligning these images with extreme accuracy is not difficult for us because we understand how scenes change during seasons, and we can use local and global cues in the image. However, to the best of our knowledge, vision algorithms have struggled to solve this task effectively due to the lack of texture and the stylistic differences. 

Modern alignment techniques focus on two approaches, sometimes combining both. The first approach is to extract keypoints such as with SIFT or HOG~\cite{TLindberg,NDalal}. Keypoints work well with high amounts of texture, can be robust to stylistic or photometric differences, and can align large geometric warps. However, keypoint systems struggle with low-texture images. Recent attempts have been made to use keypoint-like methods for matching UAV imagery to pre-captured satellite images~\cite{MShan,AYol}. However, these methods fail to show robustness in situations with low texture (non-urban environments) or across large temporal variation.

In the cases of low-texture imagery, it is often beneficial to use dense image alignment techniques. The most basic of such dense techniques is to perform the Lucas-Kanade (LK) optical flow algorithm on the raw pixels of the images. Advanced dense techniques extract a dense descriptor of the pixels before using the LK formulation~\cite{HAlismail,EAntonakos}. Dense descriptors provide invariance to photometric differences, and can be accurate on low- and high-texture scenes. However, they are limited in the magnitude of the warps that they can determine.

Verdie et al. have proposed a method which is able to extract key-points which are temporally-invariant across time of day and weather~\cite{YVerdie}. However, their testing is limited to highly-textured imagery of urban environments where there is rich information for key-point extraction. We show in the Experiments section that our method can perform well on their dataset, as well as low-texture satellite imagery. Yang et al.~\cite{KYang} present a state-of-the-art algorithm for automatic image registration of satellite imagery. Their approach combines several features to use for registration, which include standard Euclidean distance of pixel intensity, the Shape Context feature, and SIFT distance. These feature representations are all reliant on ample texture and features in the images to be aligned. Yang et al. do not show that their method can perform well on low-texture imagery, or for cases of large temporal differences.

Recently, we have seen the introduction of a technique for image alignment which uses fully-convolutional networks to extract dense features of an image and template, before using the LK formulation on the extracted features~\cite{CWang,CChang}. We henceforth refer to these methods as \tb{deep alignment methods}. The benefit of these methods is that the features relevant to alignment can be learned based on the image domain.

In this paper, we make three main contributions:

\begin{enumerate}
    \item We propose the use of deep alignment methods for aligning images of outdoor scenes that were taken across large temporal gaps, and show the viability of this approach.
    \item We explore the generalization ability of deep alignment for aligning outdoor scenes, by training on outdoor images from one domain (satellite images) and testing on images from another domain (time-lapse images taken near ground level).
    \item We propose a more optimal training strategy than has yet been proposed for deep alignment methods. Specifically, we show that using dynamic LK iterations during training provides dramatic performance benefits over single-iteration methods (see section \ref{iter_training} for more information).
\end{enumerate}

\section{Methodology}

Our approach to the image alignment problem generally follows from the approach proposed in~\cite{CChang,CWang}. The approach is to use a fully-convolutional network followed by a differentiable, iterative Lucas-Kanade implementation. The input to the Lucas-Kanade implementation is the multi-channel feature representation that is extracted from the fully-convolutional network. The entire network, from the input image to the output of the Lucas-Kanade layer, is differentiable and thus we can perform backpropagation to the determine the weights of the convolutional layer parameters. In this way, we can optimize the convolutional weights for the image alignment task. We review the selection of the convolutional network in these deep alignment architectures. We also review the fundamentals of the Inverse-Compositional Lucas-Kanade layer. For more in-depth treatment of these topics, we refer the reader to prior deep alignment work.

\subsection{Fully Convolutional Neural Network}

For the convolutional part of the network, the approaches vary between the implementations of~\cite{CChang} and~\cite{CWang}. For CLKN, they have a large enough dataset that it becomes feasible to train the convolutional weights from scratch for the image alignment task. For DeepLK, given limited training data for the object tracking task, they choose to fine-tune the conv5 layer of AlexNet. For our approach, due to a similar problem of a lack of data in the problem domain, we choose to fine-tune the conv3 layer of the VGG16 network~\cite{vgg}. We use the same convolutional weights to extract features from both the image and the template (coupled weights).

\subsection{Parameterization of the Warp Function}

The goal of image alignment is, given an image and a template, to find the parameters of a geometric warping function which describe the warping from template to image. In this paper, we study the projective warp function (homography). From~\cite{SBaker}, we have that a pixel located at $(x,y)$ in the template image, the homography warp function is commonly written using 8 parameters $\textbf{p}=(p_1,p_2,p_3,p_4,p_5,p_6,p_7,p_8)$:

\begin{align} \label{warp_eqn}
    \textbf{W(x;p)}=\frac{1}{1+p_7x+p_8y} 
    \begin{pmatrix}
        (1+p_1)x+p_3y+p5 \\
        p_2x+(1+p_4)y+p6
    \end{pmatrix}
\end{align}

\subsection{Inverse-Compositional Lucas-Kanade}

The Lucas-Kanade algorithm seeks to minimize the pixel-wise squared difference between a template and a warped version of the image. Note that the template and image can have an arbitrary number of channels; in our case, the input image and template are feature maps with 256 channels that have been outputted from the fully convolutional layers. 

In the Inverse Compositional formulation, the role of the template and the image are reversed to achieve much increased efficiency. The formulation of the Inverse Compositional Lucas-Kanade (ICLK) minimization for a template $\tb{T}$ and an image $\tb{I}$ is thus:

\begin{align} \label{min}
    \sum_{x} [\textbf{T(W(x;$\Delta$p)} - \tb{I(W(x;p))}]^2
\end{align}

Where $\Delta \textbf{p}$ indicates the change in warp parameters for a single iteration of ICLK, and $\textbf{p}$ is the result of the composition of all $\Delta \textbf{p}$ updates over all iterations so far. To solve Equation \ref{min}, we must perform a first-order Taylor expansion, and solve the resulting least-squares form. The resulting expression of the warp update is $\Delta \textbf{p}$:

\begin{align} \label{dp}
    \Delta \tb{p} = \tb{H}^{-1} \sum_{\tb{x}} \left[ \nabla \tb{T} \frac{\partial \tb{W}}{\partial \tb{p}} \right]^{T} [\tb{I(W(x;p))} - \tb{T(x)}]
\end{align}

Where $\tb{H}$ is the Hessian matrix:

\begin{align}
    \tb{H} = \sum_{\tb{x}} \left[ \nabla \tb{T} \frac{\partial \tb{W}}{\partial \tb{p}} \right]^{T} \left[ \nabla \tb{T} \frac{\partial \tb{W}}{\partial \tb{p}} \right]
\end{align}

To compute $\Delta \tb{p}$ using Equation \ref{dp}, we must compute the warp Jacobian $\frac{\partial \tb{W}}{\partial \tb{p}}$ at $\tb{(x;0)}$. For homography, the warp Jacobian can be written as:

\begin{align}
    \frac{\partial \tb{W}}{\partial \tb{p}} \tb{(x;0)} =
    \begin{pmatrix}
        x\ & y\ & 1\ & 0\ & 0\ & 0\ & -x^2\ & -xy\ \\
        0\ & 0\ & 0\ & x\ & y\ & 1\ & -xy\ & -y^2\
    \end{pmatrix}
\end{align}

Once the solution for $\Delta \tb{p}$ has been obtained via Equation \ref{dp}, it can be converted to a homography matrix, and applied via inverse composition to the current warp parameters. Specifically, if $H_\Delta$ is the homography with parameters $\Delta \tb{p}$, and $H_\tb{p}$ is the homography with parameters $\tb{p}$, then the updated parameters can be extracted from a new homography calculated by:

\begin{align}
    H_\tb{p} = H_\tb{p} H_{\Delta \tb{p}}^{-1}
\end{align}

\subsection{Iterations of Lucas-Kanade}

The Lucas-Kanade method solves a non-linear least squares problem by first-order Taylor expansion and successive iterations. Therefore, for each unique image and template pair, there is a variable number of iterations until convergence of the algorithm. The criterion for convergence is often a heuristic threshold on the magnitude of the change in warp parameters, $\Delta \tb{p}$. This is the chosen method of convergence used in~\cite{CChang}. In~\cite{CWang} however, the authors use the magnitude of the average error residual at each iteration as the stopping criterion. The error residual is calculated as $\tb{I(W(x;p))} - \tb{T(x)}$ in Equation \ref{dp}. We theorize that the choice of stopping criterion is highly dependent on the problem domain and the type of imagery. We experimented with using both the average residual method, and the magnitude of $\Delta \tb{p}$ threshold method. We found that using a heuristic threshold on the magnitude of warp parameters provides the best trade-off between accuracy of the final alignment and number of iterations.

\section{Training}

The steps for a single training iteration are as follows:

\begin{enumerate}
    \item Generate an image and template, and ground-truth warp parameters which relate the image to the template
    \item Extract feature maps of image and template using fully convolutional layers
    \item Input both image and template feature maps into Inverse-Compositional Lucas-Kanade algorithm
    \item Iterate the ICLK algorithm until convergence, indicated by the magnitude of $\Delta \tb{p}$
    \item Compute a final loss between the estimated warp parameters and the ground truth warp parameters
    \item Apply stochastic gradient updates to the convolutional filter weights in the fully convolutional layers
\end{enumerate}

\subsection{Loss Function}

A question arises of the correct loss function to use for the task of estimating homography parameters. In~\cite{CWang}, the authors use the Conditional Loss, which is a Huber loss computed on the difference between the ground truth warp parameters and the estimated parameters:

\begin{align}
    \sum_D \mathcal{H}(\tb{p} - \tb{p}_{gt})
\end{align}

Where $D$ is the set of all training data, and $\mathcal{H}$ indicates the Huber loss. The problem with this approach is that each of the 8 parameters within $\tb{p}$ do not equally affect the magnitude of the geometric warp. For instance, the projective parameters $p_7$ and $p_8$ carry much more weight in terms of the visual effect of the warp, than do the translational terms $p_3$ and $p_6$. Therefore, a loss function is needed which captures the visual correctness of the regressed parameters $\tb{p}$.

We use the Corner Loss proposed in~\cite{CChang}, which is a much better measure of visual correctness than the Conditional Loss. The Corner Loss computes the squared distance between the four corners of a ground truth un-warped image $\tb{I(W(x;p}_{gt}\tb{))}$ and the prediction of the un-warped image $\tb{I(W(x;p))}$. Defining the 4 corners of the warped image $\tb{I}$ as $\tb{c}_1$, $\tb{c}_2$, $\tb{c}_3$, $\tb{c}_4$, we have the Corner Loss defined as:

\begin{align} \label{cl}
    L(\tb{p}_{gt}, \tb{p}) = \sum_{i=1}^{4} || \tb{W(c}_i;\tb{p)} - \tb{W(c}_i;\tb{p}_{gt}\tb{)} ||^2_2
\end{align}

\subsection{ICLK Iterations During Training} \label{iter_training}

In both~\cite{CWang} and~\cite{CChang}, the authors only allow the ICLK layer of the network to iterate a single time during the training stage, although during testing the ICLK is able to iterate to convergence. We believe that they take this approach due to limitations of the implementation framework, which make it very difficult to perform back-propagation through multiple ICLK iterations during training. Therefore, the authors design loss functions which suit the single-iteration training regime, and augment their dataset to mimic the middle output of LK iterations. However, we use the more optimal strategy of iterating a dynamic number of times in the ICLK layer during training. Iterating dynamically during training also allows us to utilize the simple formulation of Corner Loss in Equation \ref{cl} as the loss function of our network output.

We have seen that unfolding dynamic LK iterations during training gives a dramatic performance improvement over single LK iteration during training for our task at test time. Further information can be found in Section \ref{single_vs_dynamic_sect} and Figure \ref{single_vs_dynamic_res}, where we compare single-iteration Corner Loss minimization and dynamic-iteration Corner Loss minimization.

\section{Datasets}

Training and testing our algorithm requires the gathering of a large amount of aligned images of a variety of outdoor environments, across many different periods of time which showcase seasonal changes and other temporal differences. One source of such data is aligned orthographic satellite imagery. Using such data has been common in approaches to remote sensing applications and GPS-denied UAV localization. Another source of data is the Archive of Many Outdoor Scenes (AMOS) dataset, which is also used by~\cite{YVerdie}. AMOS provides data generated via webcams positioned at many outdoor scenes across the world, effectively generating large amounts of time-lapse data of outdoor scenes.

\subsection{Satellite Imagery Dataset}

\begin{figure}
\centering
\includegraphics[height=3.8cm]{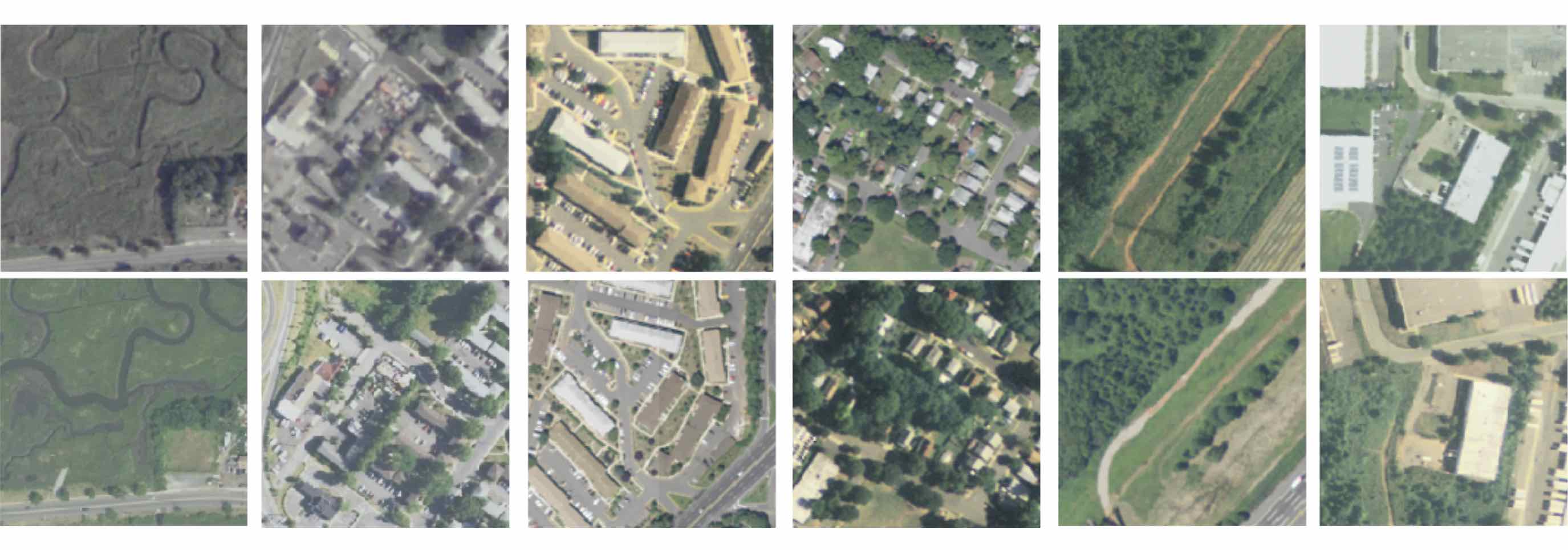}
\caption{Some representative patch examples from our \tb{New Jersey} satellite image dataset. The examples illustrate challenges of the temporally-invariant alignment problem; namely, varying levels of texture, image quality, stylistic changes, degradation in natural environments, the addition or subtraction of structures such as buildings, and shadows. \tb{We find that our algorithm is the only method of the ones we experimented with which is, in fact, able to align all of the above examples.}}
\label{new_jersey_examples}
\end{figure}

We obtained satellite imagery data from the United States Geographical Survey Earth Explorer (earthexplorer.usgs.gov) website. For training and testing, we use images from a suburban area of New Jersey, USA. We chose the location for its abundance of data, and its even mix between high and low texture. Some example images from this dataset are shown in Figure \ref{new_jersey_examples}. We obtain aligned images taken during spring, summer, and fall, taken in 2006, 2008, 2010, 2013, 2015, and 2017 (10 images total). The images are each 7582$\times$5946 pixels, at a resolution of 1 meter per pixel, meaning we train on about 50 sq. km. of geographical area. We withold 20\% of the dataset for testing.

We dynamically create data pairs from the satellite imagery data during training and testing. That is, during the training loop, we randomly choose two of the 7582$\times$5946 images from the New Jersey dataset, and randomly choose a location in the image and a patch size to sample. Keeping one of the patches static (the template $\tb{T}$), we apply a random warping to the other patch (the image $\tb{I}$). The parameters for random patch size, and the random warp parameters, can be found in the Implementation Details section.

\subsection{AMOS Time-lapse Dataset}

\begin{figure}
\centering
\includegraphics[height=3.8cm]{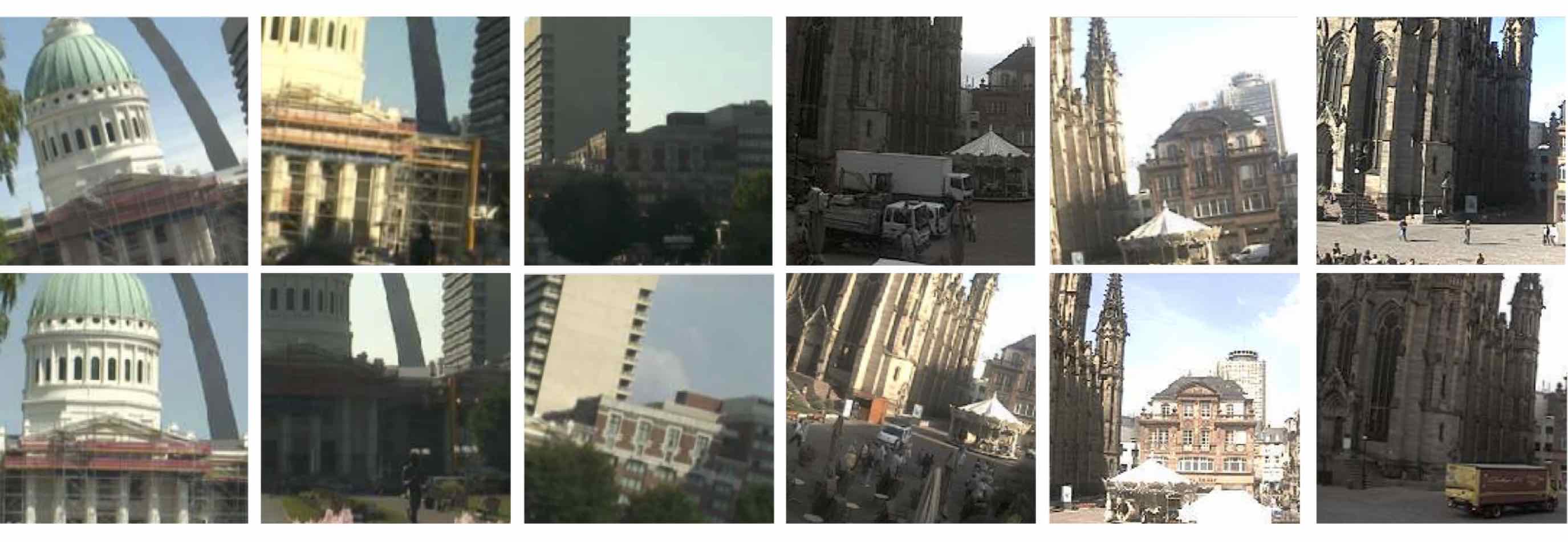}
\caption{Some representative patch examples from the time-lapse webcam dataset~\cite{YVerdie}. The three example data pairs on the left are from the \tb{StLouis} testing dataset, and the three example pairs on the right are from the \tb{Courbevoie} dataset. We show that invariance that is learned from the satellite data can be transferred to the webcam alignment task; our algorithm performs the best among methods we experimented with on aligning the above examples, despite training only on satellite data.}
\label{web_cam_examples}
\end{figure}

In~\cite{YVerdie}, the authors have constructed a representative dataset from some of the higher quality time-lapse webcams in the AMOS dataset. Some examples from this dataset are shown in Figure \ref{web_cam_examples}. The AMOS dataset is characterized by high-texture urban environments that are very rich with features. This is very different from the satellite imagery dataset, which is characterized by lower texture and lower resolution images that may not have salient features which can be easily extracted with sparse keypoint methods. The satellite imagery also has a larger number of examples of natural scenes, such as foliage, forests, and rivers, which are almost always more challenging for the temporally-invariant alignment problem. Similarly to the satellite imagery dataset, we create data pairs dynamically from the AMOS dataset during testing.

\section{Experiments}

Our goal is to show that it is possible to learn temporal invariance for outdoor scenes in general. Thus, we propose to train our network using the satellite imagery dataset \tb{only}, and test performance on both satellite imagery and the AMOS dataset. We hypothesize that by training on satellite imagery, we can learn invariances for both low and high texture, across the paradigms of most outdoors scenes (urban, suburban, rural), and across the variations that occur from large temporal differences in the images. We show in the following experiments that this is indeed possible.

We compare our alignment strategy against two other dense alignment strategies, and one sparse key-point strategy. These strategies are:

\begin{enumerate}
    \item \tb{ICLK}: We perform the ICLK algorithm on the raw image pixels, without applying any dense descriptors. We found that normalizing the pixels to have zero-mean and unit variance per-channel improved performance for normal ICLK.
    \item \tb{ICLK on untrained VGG16 conv3 features}: We perform the ICLK algorithm on feature maps extracted from the stock VGG16 conv3.
    \item \tb{SIFT+RANSAC}: We extract sparse SIFT keypoints and perform RANSAC to estimate the homography for alignment. We use the implementation included in OpenCV. It should be noted that in~\cite{CChang}, the authors find that SIFT+RANSAC provides the second best alignment method, behind only their implementation.
\end{enumerate}

We employ a Corner Error metric for showing performance of our algorithm, which is similar to that of~\cite{CChang}. The Corner Error is related to the Corner Loss, except that it reports the average Euclidean distance between the four corners of the ground truth un-warped image $\tb{I(W(x;p}_{gt}\tb{))}$ and the prediction of the un-warped image $\tb{I(W(x;p))}$. It is measured in pixels, and is equal to:

\begin{align} \label{corn_err}
    E(\tb{p}_{gt}, \tb{p}) = \frac{1}{4} \sum_{i=1}^{4} || \tb{W(c}_i;\tb{p)} - \tb{W(c}_i;\tb{p}_{gt}\tb{)} ||_2
\end{align}

\begin{figure}
\centering
\includegraphics[height=2cm]{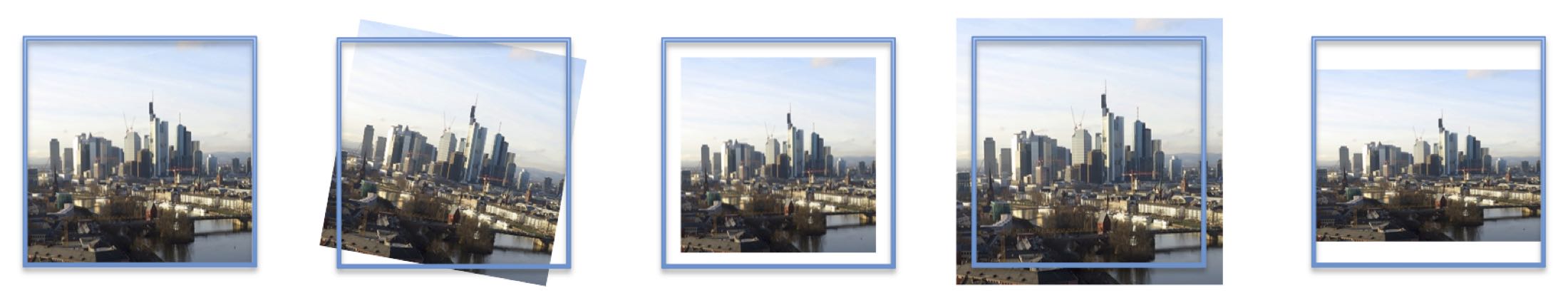}
\caption{A visualization of Corner Error. The far left picture is ground truth. The rest of the pictures illustrate warps which result in a Corner Error of 10\% of the image width from the ground truth.}
\label{corner_error_ill}
\end{figure}

Since we test with variable sizes of square image patches, we instead report the Corner Error as a percentage of the image width so that we can fairly compare warps for different image patch sizes. We provide a visualization of Corner Error in Figure \ref{corner_error_ill}.

\subsection{Implementation Details}

For generating image pairs, we randomly select two aligned images from a given dataset. We extract square patches in the images which range from 175 pixels to 300 pixels wide for the satellite image dataset, and square patches between 150 pixels and 220 pixels wide from the AMOS dataset. We extract a padded version of the image $\tb{I}$, so that when it is warped, there are not cutoff regions around the edges. We warp the image $\tb{I}$, choosing projective warp parameters from uniform random distributions. We choose warp parameters such that if the algorithm were to predict $\tb{p}=0$ for every test example (no-op), the maximum Corner Error would be about 30\% of the image width.

We transfer the conv3 layer of the VGG16 network for our convolutional pipeline, and fine-tune only conv3. We implement the algorithm using the open-source PyTorch framework, on an NVIDIA GeForce Titan X GPU. We trained on 15,000 dynamically generated training pairs from the New Jersey satellite image dataset. We implemented a mini-batch training approach, calculating the Corner Loss on a mini-batch of 5 training pairs before applying the stochastic gradient update. We generate a validation batch of 20 image pairs randomly at train time from the source data (\tb{New Jersey} satellite data), and keep the model which generates the lowest validation loss during training.

\subsection{Evaluation on Satellite Dataset}

\begin{figure}
\centering
\includegraphics[height=4.5cm]{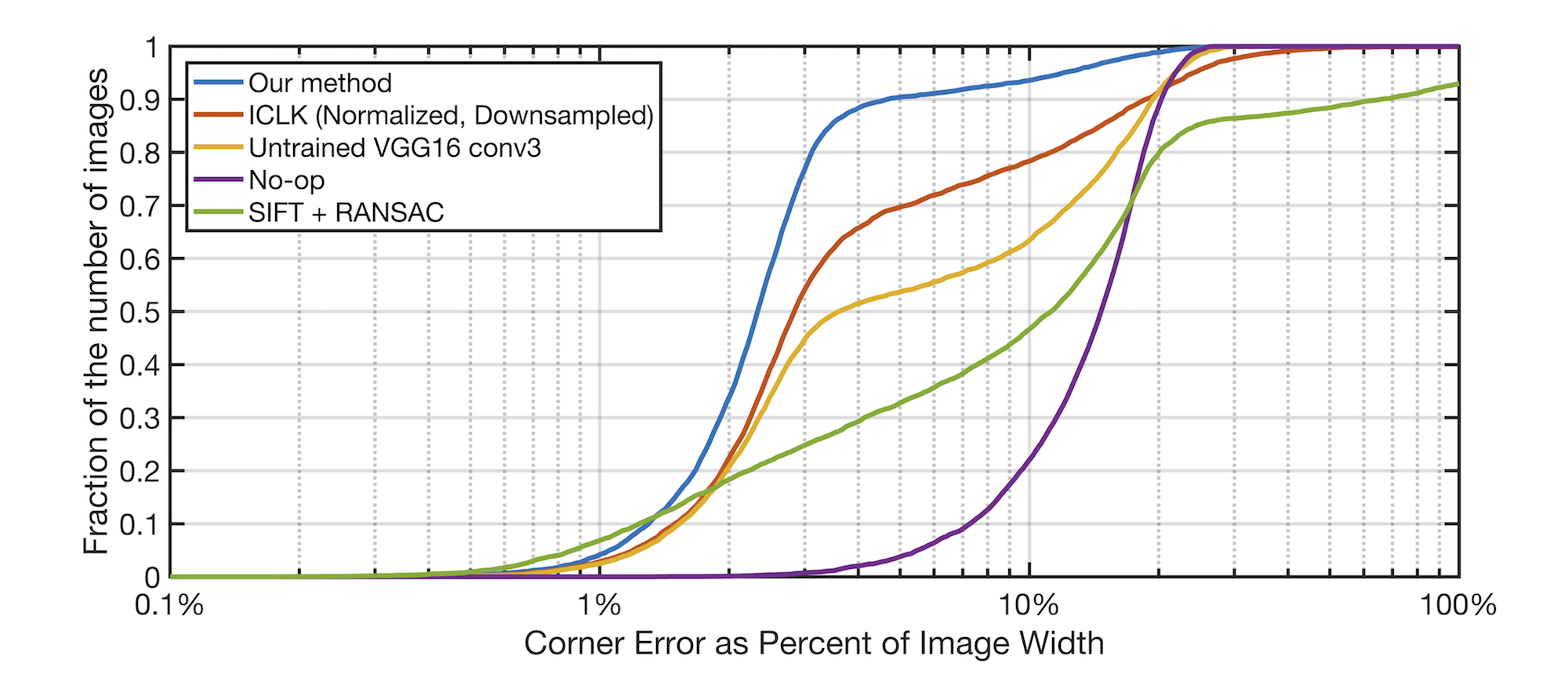}
\caption{Results of testing on 5000 test patches from our \tb{New Jersey} satellite image dataset. The Corner Error as a percentage of image width is reported. The results compare our method (blue), normal ICLK, ICLK computed on untrained VGG16 conv3 features, and SIFT+RANSAC. There are several interesting aspects of the results; for one, we can see that SIFT+RANSAC can align about 7\% of the dataset to less than 1\% Corner Error. However, at the 1.5\% Corner Error threshold, we see that our method (blue) rapidly surpasses the performance of SIFT+RANSAC and the other methods. Our method aligns 80\% of the dataset to less than 3\% Corner Error. Purple illustrates the results for an algorithm which always returns $\tb{p}=0$ (no-op). We can see that all methods except ours (blue) pass through the purple line, indicating that they perform worse than no-op for some ratio of the dataset.}
\label{new_jersey_result}
\end{figure}

\begin{figure}
\centering
\includegraphics[height=7.15cm]{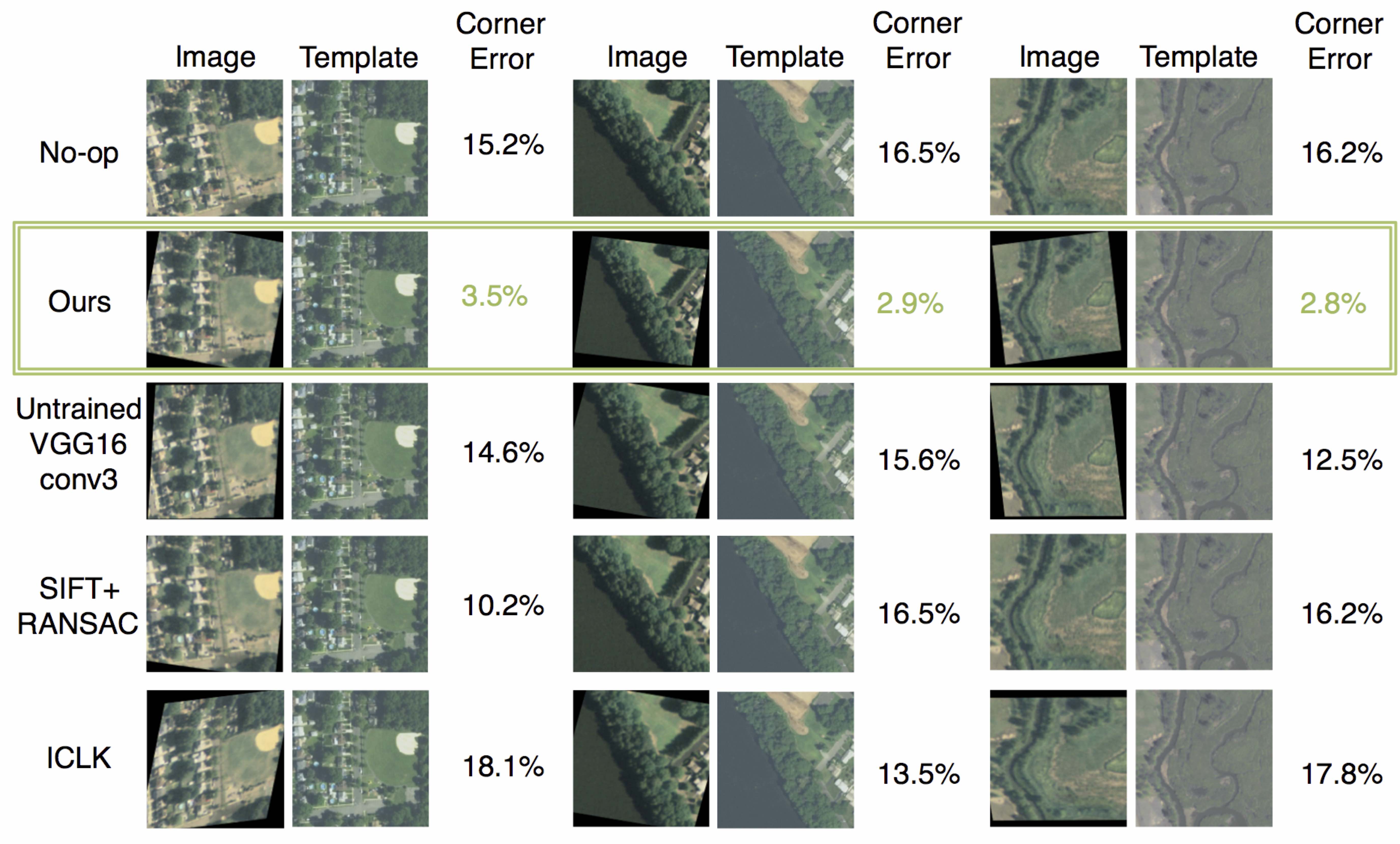}
\caption{Notable results from testing on the \tb{New Jersey} satellite image dataset. Our method has the ability to perform alignment on low texture scenes, despite significant style differences due to temporal effects. In many experiments, the other methods have final Corner Error that is greater than the Corner Error of no-op. We also find that SIFT+RANSAC is unable to estimate any warp at all due to lack of texture in some cases.}
\label{new_jersey_notable}
\end{figure}

We test on 5000 data pairs from the \tb{New Jersey} satellite image dataset which are unseen during training. The results of this experiment are shown in Figure \ref{new_jersey_result}. We report the results in terms of Corner Error (Equation \ref{corn_err}) as a percent of image width, versus the ratio of training data. The results indicate the superior performance of our method for aligning satellite imagery, in the face of large temporal and seasonal variations. Please see the description in Figure \ref{new_jersey_result} for more information on the performance metrics. In Figure \ref{new_jersey_notable}, we provide some notable alignment results achieved by our method.

\subsection{Evaluation on AMOS Dataset}

\begin{figure}
\centering
\includegraphics[height=7.15cm]{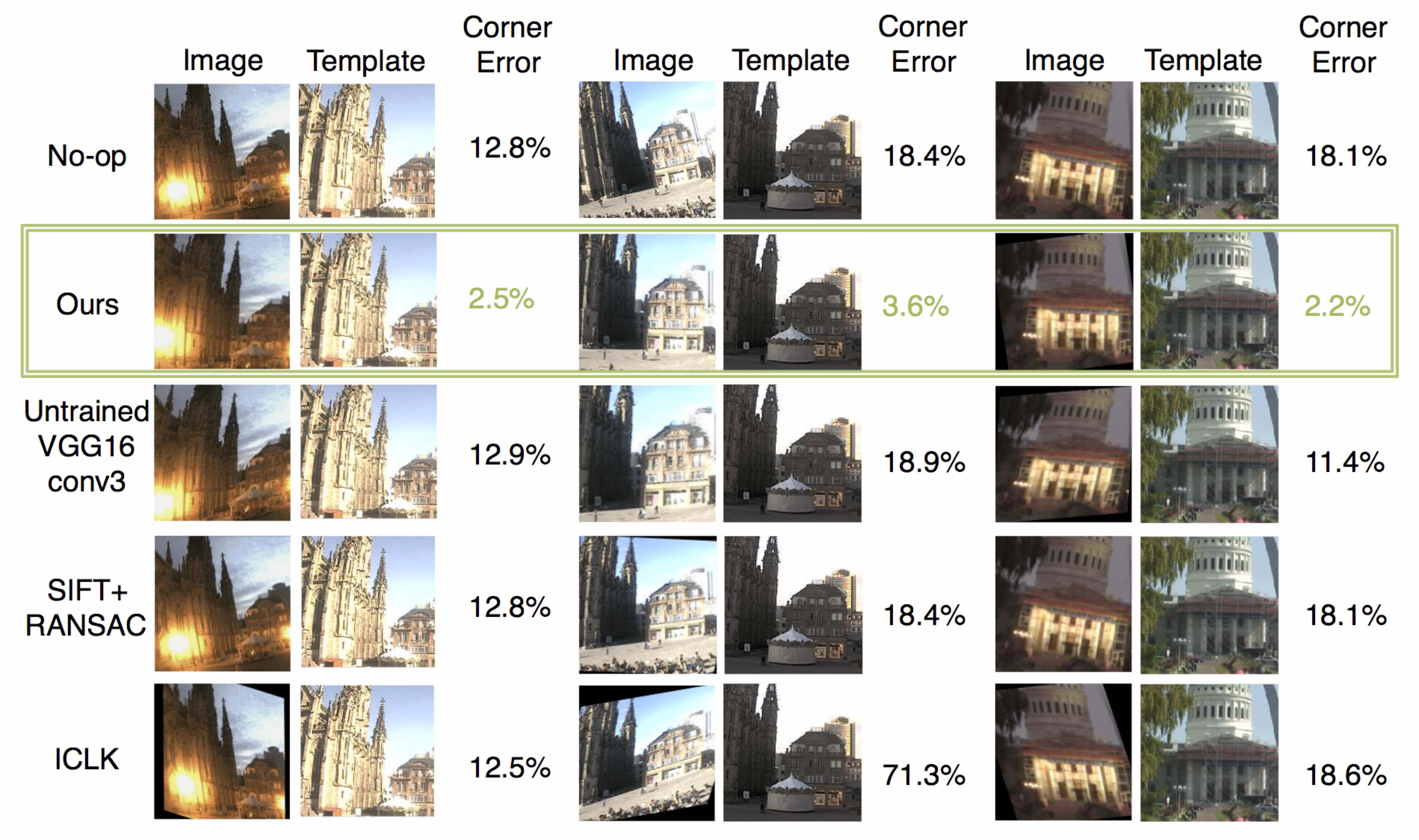}
\caption{Notable alignment results using our algorithm to align time-lapse AMOS data, using features trained only on satellite imagery. These results show that deep features can be used to learn temporal-invariance in one domain (satellite imagery) and transfer this invariance to a similar but unseen domain (time-lapse imagery from near ground-level).}
\label{webcam_notable}
\end{figure}

\begin{figure}
\centering
\includegraphics[height=5cm]{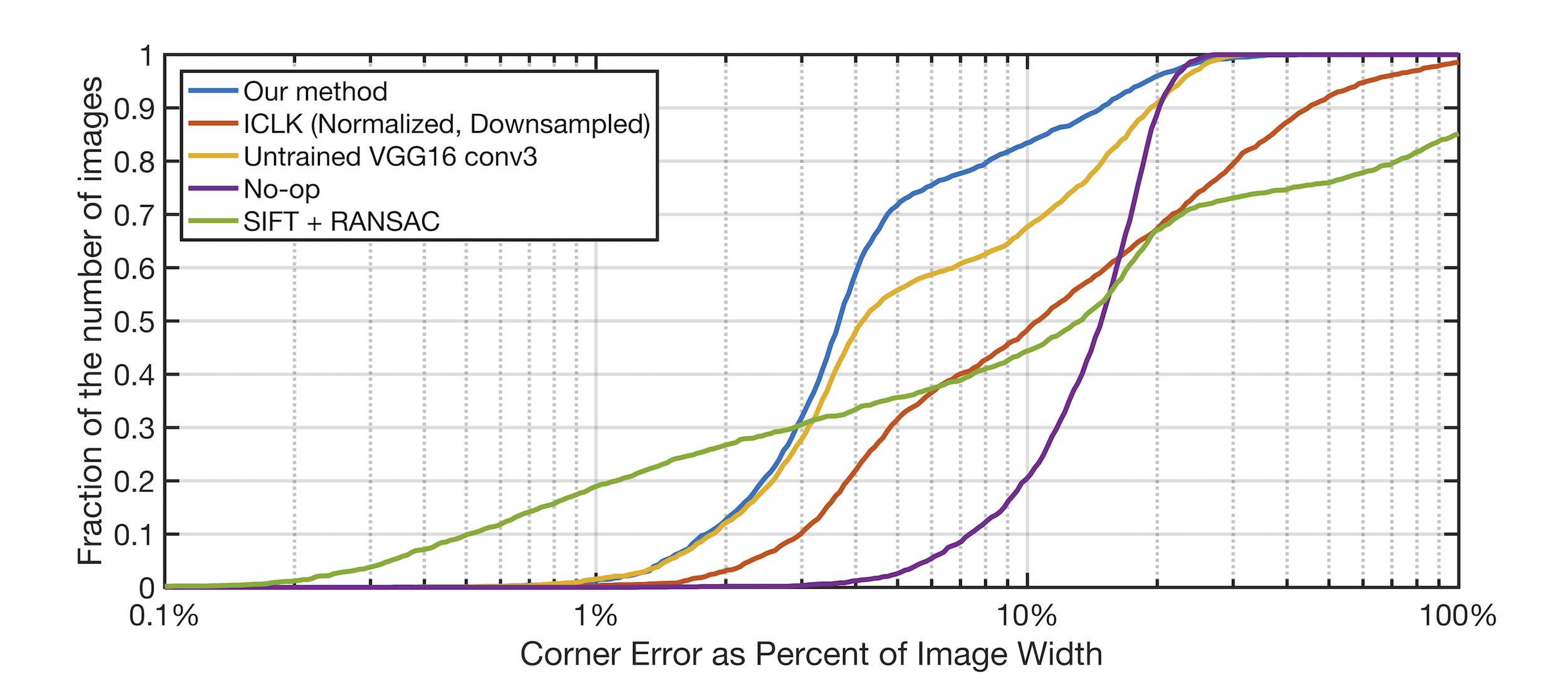}
\caption{Results for Corner Error when testing on the \tb{StLouis} webcam dataset, after training only on satellite images. We can see that our method is able to align 70\% of the test pairs to less than 5\% Corner Error. This indicates that we have learned invariance to outdoor scenes from satellite data, and have transferred that invariance to the task of webcam data alignment.}
\label{stlouis_results}
\end{figure}

For the AMOS Dataset, we first test on 2000 data pairs from a webcam located in St. Louis, Missouri, USA. Some representative examples of this dataset are shown in Figure \ref{web_cam_examples}. The alignment results for this dataset are shown in Figure \ref{stlouis_results}. Notably, we find that our alignment method, which has been trained only on satellite data, has learned invariances to outdoor scenes which allow it to be effective at alignment on the AMOS dataset. Some alignment examples are captured in Figure \ref{webcam_notable}.

\begin{figure}
\centering
\includegraphics[height=5cm]{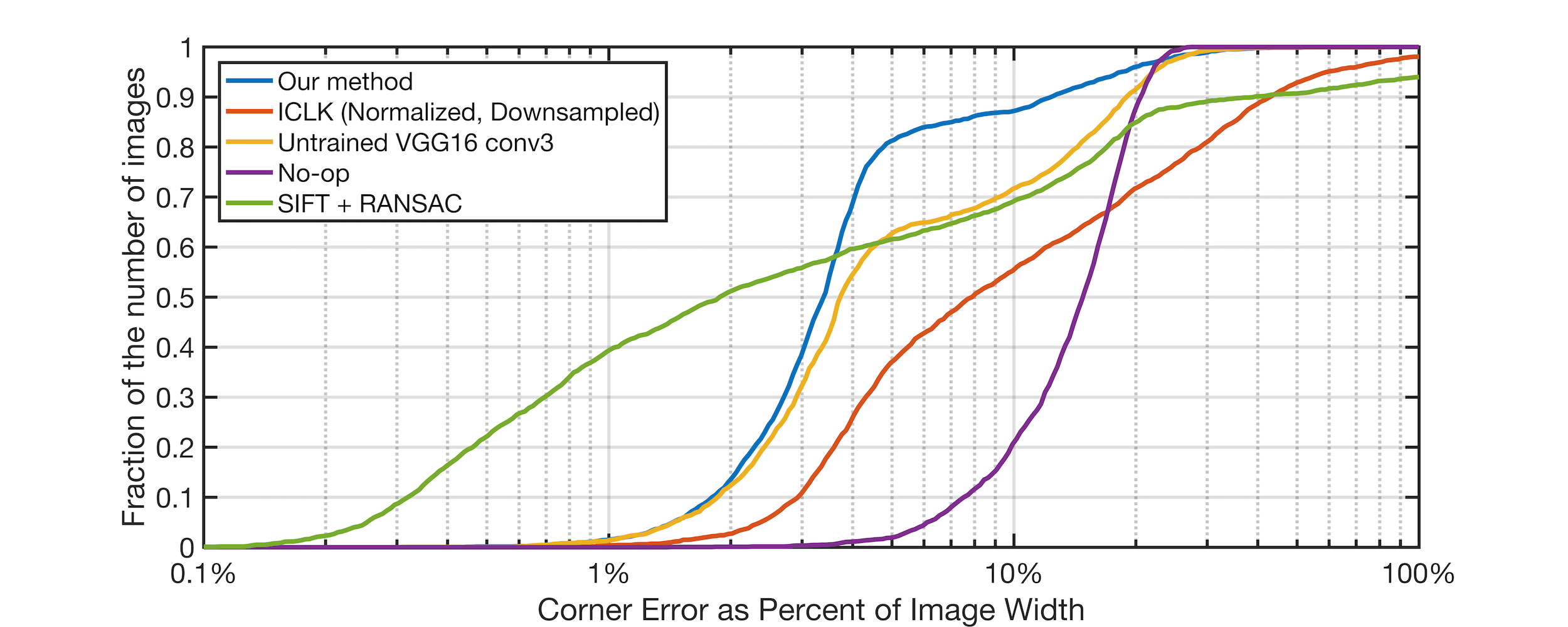}
\caption{Results for Corner Error when testing on the \tb{Courbevoie} time-lapse webcam dataset, after training only on satellite images. Although SIFT+RANSAC performs well because of the high-texture scenes, our method our method can align 80\% of the test pairs to less than 5\% Corner Error, the best of all other methods at the 5\% Corner Error threshold. The invariance that is learned by training the conv3 layer on satellite images is persistent when testing on the unseen data; we can see this from the fact that our method outperforms the untrained VGG16 conv3 layer.}
\label{courbevoie_results}
\end{figure}

We test also on 2000 data pairs from a webcam located in Courbevoie, France. Some representative examples of this dataset are shown in Figure \ref{web_cam_examples}, with the alignment results show in Figure \ref{courbevoie_results}. Again, we find that the network trained only on satellite images is able to generalize for this alignment task. Specifically, we find that our network can align 80\% of image pairs in the Courbevoie dataset to within 5\% corner error. The SIFT+RANSAC method, and ICLK on untrained VGG16 conv3 features, both align about 60\% of this dataset to within 5\% corner error.

\subsection{Single-Iteration vs. Dynamic-Iteration Training} \label{single_vs_dynamic_sect}

\begin{figure}
\centering
\includegraphics[height=5cm]{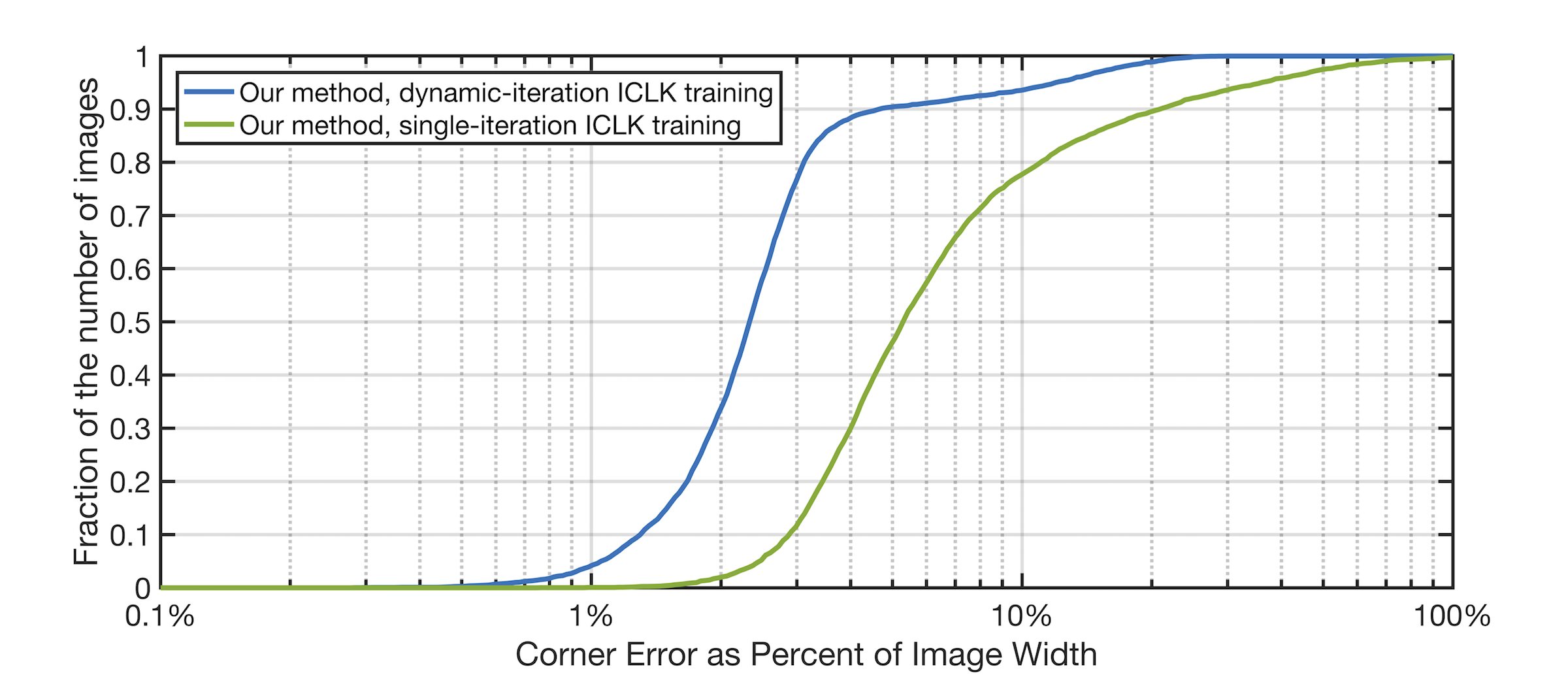}
\caption{Performance difference for single-iteration training methods of deep alignment on satellite imagery, vs dynamic iterations. The testing data is identical to the testing data shown in Figure \ref{new_jersey_result}. The single-iteration trained network has much higher Corner Error percentage throughout the testing dataset. We attempted training several times using both single-iteration and dynamic-iteration, and present the best results for both.}
\label{single_vs_dynamic_res}
\end{figure}

There is a choice of how many iterations to do in the ICLK layer of the deep alignment methods. Previous works have designed the system around a single-iteration training scheme due to limitations of the implementation framework. However, we show that dynamic iterations provide a dramatic performance boost for our task in Figure \ref{single_vs_dynamic_res}.

\section{Conclusion}

In this paper we present a new perspective on recent deep alignment techniques, applying them to the problem of temporally-invariant image alignment for outdoor scenes. We have shown a surprising real-world application of these deep alignment methods, as well as introducing a more optimal training strategy for learning temporal invariance. We have shown that, with deep alignment methods, we can learn invariances in one domain and transfer that invariance to a separate and unseen domain. We propose that this method can be used in future work in the remote sensing community, in the localization of UAVs in GPS-denied environments, 3D reconstruction algorithms, SLAM, and in other applications where high levels of invariance to imaging conditions are required.
\clearpage

\bibliographystyle{splncs}
\bibliography{eccv2018submission}
\end{document}